\documentclass[10pt,twocolumn,letterpaper]{article}
 \usepackage{cvpr}
\usepackage{times}
\usepackage{epsfig}
\usepackage{graphicx}
\usepackage{amsmath}
\usepackage{amssymb}
\usepackage{booktabs}
\usepackage{mathrsfs}
\usepackage{lipsum}
\usepackage{mathtools}
\usepackage{stackrel}
\usepackage{cuted}
\include{gdefs}
\newcommand{\flf}{\hat{f}^{\text{LF}}}
\newcommand{\fhf}{\hat{f}^{\text{HF}}}

\usepackage[pagebackref=true,breaklinks=true,letterpaper=true,colorlinks,bookmarks=false]{hyperref}

\cvprfinalcopy 

\def\cvprPaperID{6174} 

\ifcvprfinal\fi
\begin{document}

\title{Learning to synthesize: splitting and recombining low and high spatial frequencies for image recovery}

\author{Mo Deng\thanks{These authors contributed equally}\\
{\tt\small modeng@mit.edu}
\and
Shuai Li\footnotemark[1]\\
{\tt\small shuaili@mit.edu}
\and
George Barbastathis\\
{\tt\small gbarb@mit.edu}
}

\maketitle

\begin{abstract}
Deep Neural Network (DNN)-based image reconstruction, despite many successes, often exhibits uneven fidelity between high and low spatial frequency bands. In this paper we propose the Learning Synthesis by DNN (LS-DNN) approach where two DNNs process the low and high spatial frequencies, respectively, and, improving over \cite{Pan_2018_CVPR}, the two DNNs are trained separately and a third DNN combines them into an image with high fidelity at all bands. We demonstrate LS-DNN in two canonical inverse problems: super-resolution (SR) in diffraction-limited imaging (DLI), and quantitative phase retrieval (QPR). Our results also show comparable or improved performance over perceptual-loss based SR \cite{johnson2016perceptual}, and can be generalized to a wider range of image recovery problems. 
\end{abstract}

\section{Introduction}

The noiseless forward model of a computational imaging system is
\begin{equation}
g=Hf, 
\label{eq:gHf}
\end{equation}
where $f$ is the unknown object, $H$ is the forward operator and $g$ is the raw intensity image captured by the camera. The standard method of solving this inverse problem for linear ill-posed forward operators $H$ is 
\begin{equation}
   \hat{f}=\stackrel[f]{}{\text{argmin}} \left\{\rule[-1ex]{0cm}{3ex} 
\left|\!\left| Hf-g \right|\!\right|^2 +
\Phi(f)
\right\},
\label{eq:minfunc}
\end{equation}
where $\Phi(f)$ expresses prior knowledge about the class of objects of interest, and is known as the regularizer. Compressive imaging techniques\cite{inv:candes-romberg-tao06a,inv:candes-romberg-tao06b,inv:candes-tao06,inv:donoho2006compressed}, in particular, have been very effective at improving the condition of the inverse problem by enforcing sparsity constraints, as long as the basis where the objects are sparse is known. Alternatively, in dictionary-based methods \cite{inv:elad2006image,inv:aharon2006k,inv:rubinstein2010dictionaries} the basis set is {\em learnt} from examples. If more general priors are applicable to the problem, then machine learning techniques, especially Deep Neural Networks \cite{lecun2015deep} (DNNs) become attractive. 

Several forms of machine learning architectures have been used for solving problems of the form (\ref{eq:minfunc}). One approach inspired by the Learning Iterative Shrinkage and Thresholding Algorithm (LISTA) \cite{gregor2010learning}, is a cascade of DNNs with the conjugate forward operator $H^t$ acting as residual \cite{inv:mardani2017b}. Two simpler alternatives are the end-to-end approach, where the input to the DNN is directly the raw intensity image $g$ and the output is the estimate $\hat{f}$\cite{sinha2017lensless,rivenson2017deep}; and the preprocessor approach, where $g$ is fed into an approximant $H^*$ before going into the DNN \cite{rivenson2017deep,goy2018low}. The quality of the inverse estimate $\hat{f}$ depends on our knowledge of the prior $\Phi$ if (\ref{eq:minfunc}) is used; and on the content of the training example database if a machine learning approach is used. 

One feature of the examples that we have found to strongly influence training is the examples' spatial frequency content relative to the spatial frequency behavior of the forward operator $H$. Let us consider the case when $H$ suppresses high frequencies, which is often encountered in practice due to undersampling or the finite aperture of optical systems; and that we train using a database such as ImageNet, which is well-known to have an inverse-square law (spatial) power spectral density (PSD) \cite{van1996modelling}. In that case, does the DNN learn the inverse square law? Li \textit{et al} have found \cite{li2018spectral} that this is not necessarily the case. Because high frequencies are sparser in the database, and the training process is highly nonlinear, low frequencies may end up dominating the prior by more than their fair share; low-pass filtering of the inverse $\hat{f}$ and loss of fine detail ensues. The same work \cite{li2018spectral} proposed a spectral pre-modulation approach to compensate for the scarcity of high spatial frequencies in the database and showed that indeed fine details are recovered; however, at the same time, high-frequency artifacts appeared in the reconstructions, evidently because the spectral pre-modulation also taught the DNN to edge enhance. 

In this paper, we propose a novel Learning Synthesis by DNN (\textbf{LS-DNN}) method to effectively manage and synthesize different spectral bands according to their relative behavior in the forward operator and prevalence in the training database. Our approach, similar to \cite{Pan_2018_CVPR}, processes the signal separately into two DNNs, assigned to low- and high-frequency bands, respectively. However, unlike \cite{Pan_2018_CVPR} we {\em learn} how to synthesize the two bands adaptively; we have also modified the training procedure. More specifically, our main contributions are:
\begin{itemize}
    \item We train the two DNNs as follows: the low-band DNN (DNN-L) is trained so that its output $\flf$ match the unfiltered examples; whereas the high-band DNN (DNN-H) is trained such that its output $\fhf$ match a filtered version of the examples where high spatial frequencies have been amplified. This choice for the DNN-L is for databases such as ImageNet, where high frequencies are naturally under-represented, as pointed out earlier. 
    
    \item We use a third DNN (DNN-S) to synthesize the reconstructions $\flf$ and $\fhf$ from DNN-L and DNN-H, respectively, into a final estimate $\hat{f}$. This ensures that the low and high spatial frequencies are rebalanced correctly in the final reconstruction, according to the relative behavior of the two bands in the forward operator and the prior expressed by the examples. 
    
    \item We demonstrate the effectiveness of the \textbf{LS-DNN} method in two canonical ill-posed problems. The first problem is super-resolution (\textbf{SR}) in diffraction-limited imaging (\textbf{DLI}); {\it i.e.}, achieving a resolving ability beyond the diffraction-limited resolution, which is effectively deblurring in the particularly severe case where the spatial spectrum of the forward operator has a hard cut-off. Since spatial frequencies are entirely eliminated from the raw measurement $g$ in this case, the only way to recover them is through the learned prior. The second problem is lensless quantitative phase retrieval (\textbf{QPR}); {\it i.e.}, the recovery of the phase of a coherent electromagnetic field from intensity measurements. This problem is subject to both high-frequency cut-off due to the system aperture and scrambling of the phases of the remaining spatial frequency components due to Fresnel propagation. 
    
    \item We demonstrate that reconstructions according to \textbf{LS-DNN} are visually comparable and sometimes better than those obtained by the widely used VGG16\cite{simonyan2014very}-based perceptual loss\cite{johnson2016perceptual} SR strategy, in the SR/DLI problem. 
    
    \item We also applied {\bf LS-DNN} to the SR problem according to \cite{Pan_2018_CVPR} and obtained reconstructions that are sharper than those in \cite{Pan_2018_CVPR}. 
\end{itemize}

\section{Related Works}
\begin{itemize}
\item \textbf{Image Super-resolution (SR)}
The term super-resolution (SR) can have different interpretations in different research communities: it can mean \textbf{upsampling} \cite{dong2014learning}, {\it i.e.} starting from the raw image $g$ captured on a limited number of pixels and arriving at an estimate $\hat{f}$, with a larger number of pixels within the same spatial support; it can also be interpreted as \textbf{deblurring}, where the blur kernel is a low-pass filter with transfer function equal to zero for spatial frequencies above cutoff \cite{goodman2005introduction}. 

An extensive review on non-machine learning based methods of single-image SR (under the upsampling interpretation), is given in \cite{yang2014single}. To our knowledge, \cite{dong2014learning} is the first effort to use DNN in this context. \cite{johnson2016perceptual,ledig2017photo} are among the important follow-ups, where visually more appealing reconstructions are made possible with the use of perceptual loss\cite{johnson2016perceptual} and Generative Adversarial Networks(GAN)\cite{goodfellow2014generative}.

To our knowledge, the only prior work using DNNs on deblurring with a hard-cutoff in frequency imposed by the kernel was \cite{rivenson2017deep}, where the low-resolution image and high-resolution image within a training pair were captured using two lenses with different numerical apertures (NAs). 

\item \textbf{Quantitative phase retrieval (QPR)}
The phase change induced when light is transmitted through an object carries crucial information about the sample. However, conventional cameras are only sensitive to the intensity rather than the phase of the light. Therefore, the task of QPR is to recover the phase information of the object from the raw intensity measurements. 
Traditional approaches for QPR include the Gerchberg-Saxton (GS) algorithm \cite{gerchberg1972practical}, digital holography (DH) \cite{goodman1967digital}, phase shifting interferometry \cite{creath1985phase}, and Transport of Intensity Equation (TIE) \cite{teague1983deterministic}. 

Deep learning was introduced for QPR by Sinha \textit{et al} \cite{sinha2017lensless} using an end-to-end residual convolutional neural network\cite{he2016deep}. Subsequently, Rivenson \textit{et al} \cite{rivenson2018phase} proposed a QPR in a DH implementation with a DNN including an approximant: first  an approximate phase reconstruction was obtained using an optical back-propagation method; that reconstruction was then fed into the neural network for further improvement. Goy \textit{et al} \cite{goy2018low} further showed that the GS algorithm is a good approximant for the DNN to perform QPR under conditions of low photon flux.

\item \textbf{Perceptual Quality Enhancement} 
Prior to \cite{li2018spectral}, it was realized that low-level pixel-wise losses generally encourage finding pixel-wise averages of plausible solutions which tend to be over-smooth\cite{gupta2011modified,wang2004image,wang2003multiscale,bruna2015super,mathieu2015deep,dosovitskiy2016generating}. Instead, perceptual loss, based on high-level image feature representations extracted from pre-trained CNNs, can be used as the loss function, generating good quality images. This strategy was applied to feature inversion by Mahendran \textit{et al}\cite{mahendran2015understanding}, to feature visualization by Simonyan \textit{et al}\cite{simonyan2013deep} and Yosinski \textit{et al}\cite{yosinski2015understanding}, and to texture synthesis and style transfer by Gatys \textit{et al}\cite{gatys2015texture}. These approaches generally require solving an optimization problem, which can be slow. 
Johnson \textit{et al}\cite{johnson2016perceptual} is the first work that combines the benefits of the fast rendering of output by a CNN and perceptual loss to train feed-forward convolutional neural networks for style transfer and SR. Subsequently, Ledig \textit{et al}\cite{ledig2017photo} proposed a SR generative adversarial neural network\cite{goodfellow2014generative}(SRGAN), where the perceptual loss based on high-level feature maps of the VGG-networks\cite{simonyan2014very} and a discriminator, jointly encourage reconstructions to be perceptually indistinguishable from the high resolution images. 

\item \textbf{Dual frequency band reconstruction}
In 2008, HiLo microscopy \cite{lim2008wide} was proposed by Lim \textit{et al} to enhance the resolution of optical imaging system. Their idea was to capture one low frequency band image and one high frequency band image by applying different illumination patterns. Then, a fusion of these two images will lead to a full resolution image.

In 2018, Pan \textit{et al} proposed a dual convolutional neural network (DualCNN)\cite{Pan_2018_CVPR} structure consisting of two branches to solve low-level vision problems. They also used separate processing by two DNNs in two spatial frequency bands, low and high. However, two key differences from our approach are: first, that the two DNNs were trained together, empirically assigning the training loss function as a linear combination of the pixel-wise loss on the respective outputs; whereas in our approach the DNNs for low and high bands are trained separately, and the examples are pre-filtered before presentation to the DNN operating in the high band. The second difference is that in Pan the outputs of the low and high band DNNs are combined in {\em ad hoc} fashion, whereas we use a third DNN for the combination. Our addition of the third DNN is important, because the combination is learnt and, therefore, does a better job at reconstructing all frequency bands evenly and without artifacts, as our results show.  Also, in our approach, no prior knowledge on the forward model is needed.
\end{itemize}

\begin{figure*}[h]
\centering\includegraphics[width=0.7\linewidth]{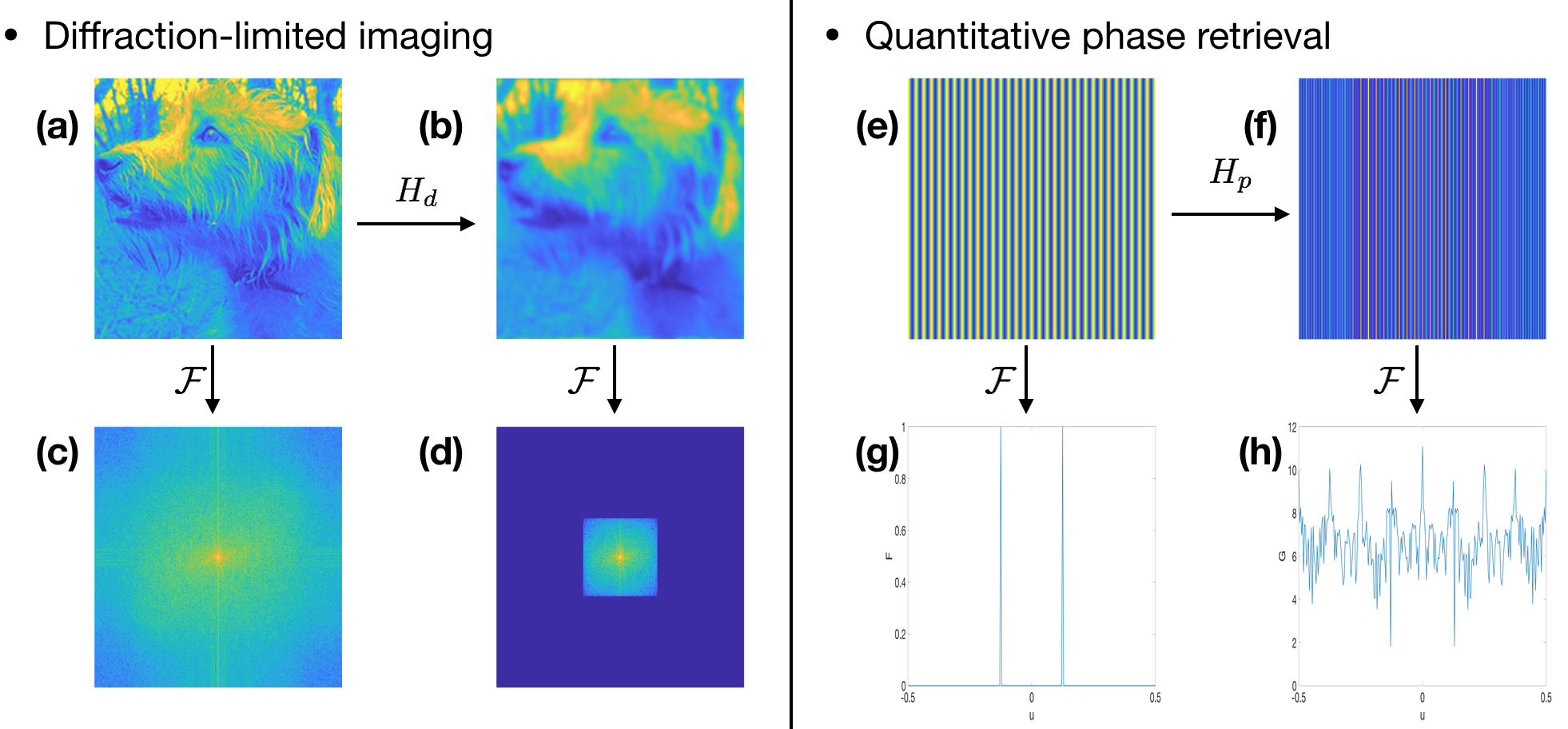}
\caption{Frequency Spectrum of measurements. (a) An ImageNet example, (b) Diffraction-limited measurement, (c) and (d) are the Fourier spectrum of (a) and (b), respectively. (e) A sinusoidal phase object , (f) Intensity measurement, (g) and (h) are 1D cross-sections along the horizontal direction for the Fourier spectrum of (e) and (h), respectively.}
\label{fig:fs}
\end{figure*}

\section{Physical Models}
 \subsection{Super-resolution (SR) in diffraction-limited imaging (DLI) under incoherent illumination}
Due to diffraction, the spatial resolution of an imaging system is limited. Assuming the nominal resolving ability of an imaging system is $b$ pixels, then for a given object $f(x,y)$, the diffraction-limited measurement $g(x,y)$ using this system under incoherent illumination will be \begin{equation}
\begin{aligned}
g(x,y) &=H_{d}f(x,y)\\
&=f(x,y)\otimes\text{sinc}^{2}\left(\frac{x}{b},\frac{y}{b}\right).\\
\end{aligned}
\label{eq:Hd}
\end{equation}
Here, $H_{d}$ is the forward operator for DLI, $\otimes$ denotes convolution and $\text{sinc}(x,y)=(\sin{(\pi x)}\cdot\sin{(\pi y)})/(\pi^{2}xy)$.

\subsection{Quantitative Phase Retrieval (QPR)}
In QPR problem, assume an unknown phase object $f(x,y)$ is at the origin $z=0$ of the optical axis. The raw intensity image $g(x,y)$ measured at a distance of $z$ to the object will be,
\begin{equation}
\begin{aligned}
g(x,y)&=H_{p}f(x,y)\\
&=\left\lvert\text{exp}[if(x,y)]\otimes \text{exp}\left[\frac{i\pi(x^{2}+y^{2})}{\lambda z}\right]\right\lvert^{2}.\\
\end{aligned}
\label{eq:Hp}
\end{equation}
Here, $H_{p}$ is the non-linear forward operator for quantitative phase imaging, $\otimes$ denotes convolution,$i$ is the imaginary unit and $\lambda$ is the illumination wavelength.

\subsection{Frequency Spectrum of Measurements}
Next, we analyze the two imaging models in the spatial frequency domain. Recall that the forward kernel in DLI is effectively a low-pass filter with finite cutoff in the frequency domain. From \ref{eq:Hd} we have
\begin{equation}
G(u,v)=F(u,v)\cdot \text{tri}(bu,bv).
\label{eq:fd}
\end{equation}
Here, $F(u,v)$ and $G(u,v)$ are the Fourier transforms of $f(x,y)$ and $g(x,y)$, respectively; $u$ and $v$ denote the 2D spatial frequencies; and $\text{tri}(u,v)$ denotes the two dimensional triangular function. As shown in Fig.\ref{fig:fs}, high frequency information beyond cutoff is lost, making the forward operator $H_{d}$ ill-posed. Here we considered DLI only; generally, the situation becomes worse due to aberrations and other imperfections in the optical system.

Because of the nonlinearity in QPR, it may be thought of as a self-convolution in the frequency domain. To demonstrate this, we show one example in Fig.\ref{fig:fs}. We assume the phase object to be a sinusoidal grating, which contains only a single frequency. However, more than one frequency components exist in the measurement, as can be observed in Fig.\ref{fig:fs} (h). This frequency scrambling, together with the loss of spatial frequencies due to the finite numerical aperture make the forward operator $H_{p}$ ill-posed.

\begin{figure*}[h]
\centering\includegraphics[width=0.8\linewidth]{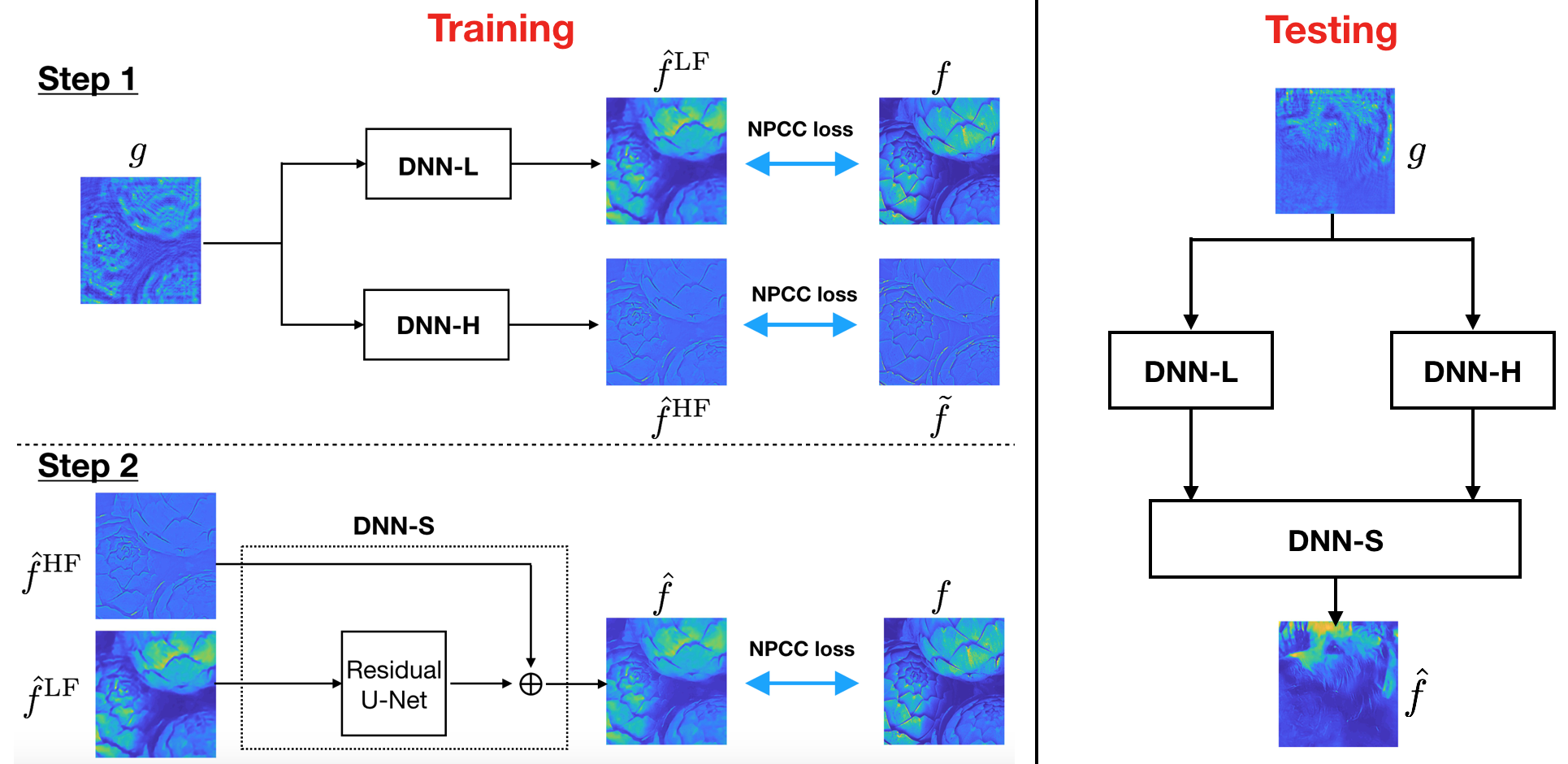}
\caption{Proposed LS-DNN.}
\label{fig:dnn1}
\end{figure*}

\section{Proposed Method}
\subsection{Learning Synthesis by Deep Neural Networks (LS-DNN)}
We propose a two-step approach (Fig.\ref{fig:dnn1}) to tackle the difficulty in restoring high frequency components. First, we use two separate deep neural networks (DNNs)  parallel,DNN-L and DNN-H. DNN-L is what typically exists in previous works -- it learns to map from the measurements to the un-filtered ground truth images; while DNN-H takes in measurements and maps them to a frequency-modulated ground truth. The modulation in the frequency domain (see Section \ref{sec:modu}), amplifies the relative weights of the high spatial frequencies, thereby facilitating the extraction of high frequencies from the priors.

We expect the reconstruction by DNN-L to be reliable in the low frequency range but not in the high frequency range; whereas the one by DNN-H should be better in the high-frequency range but distorting the low frequency range and may be creating hallucination artifacts. The training of DNN-L and DNN-H can be done in parallel. We denote the reconstructions by DNN-L and DNN-H, during  training, as $\hat{f}^{LF}$ and $\hat{f}^{HF}$, respectively.

Moreover, we train a third DNN, denoted as DNN-S, which learns to synthesize the low frequency and high frequency reconstructions from the previous networks $\hat{f}^{LF}$ and $\hat{f}^{HF}$, to generate the final reconstruction $\hat{f}$ matching the un-modulated ground truth. 

\subsection{Spectral Pre-modulation}
\label{sec:modu}
\begin{figure}[h!]
\centering\includegraphics[width=1\linewidth]{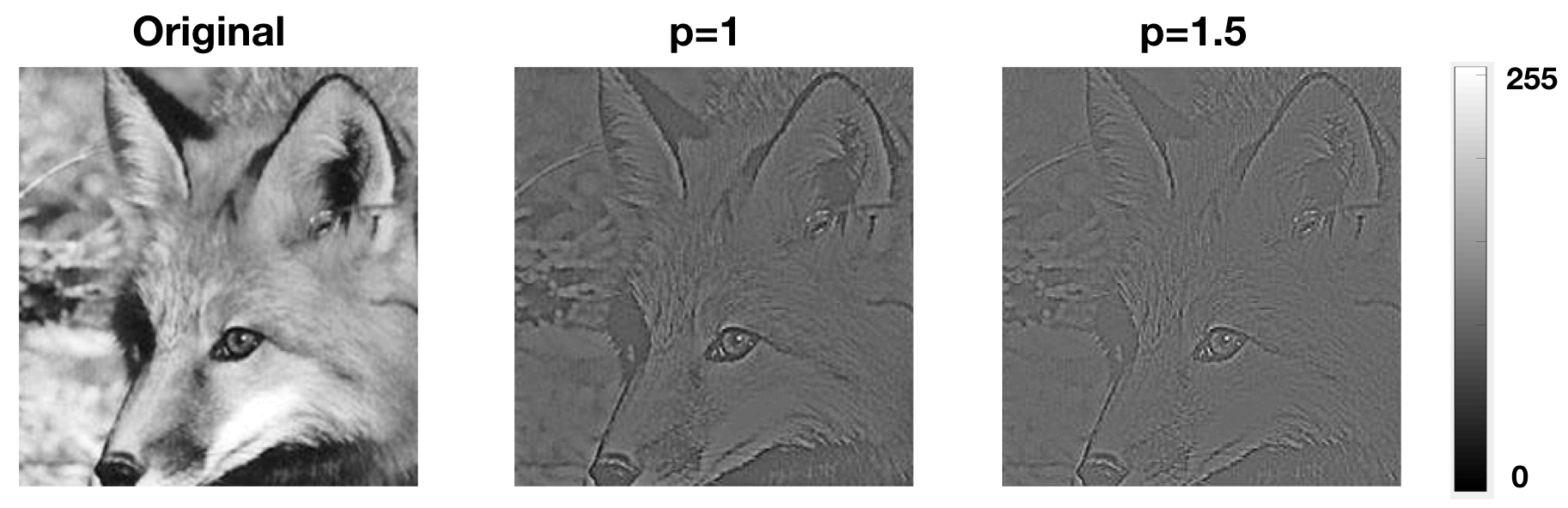}
\caption{Spectral pre-modulation results.}
\label{fig:spm}
\end{figure}

The 2D PSD of ImageNet database is well represented as
$S(u,v)\propto (\sqrt{u^2+v^2})^{-2}$. 
Therefore, we pre-modulate the ImageNet images with spatial filter $T(u,v)=(\sqrt{u^2+v^2})^{p}$, The modulated ground truth $\Tilde{f}$ has Fourier transform $\tilde{F}(u,v)=F(u,v)T(u,v)$, where $F(u,v)$ denotes the Fourier transform of an un-modulated ground truth image and $p$ controls whether high or low frequencies should be enhanced. The optimal value of $p$ is determined empirically and we find that DNN-S is robust enough to the choice of $p$ without perceptible difference in the reconstructions. In the results shown below in Section \ref{sec:result}, we use $p=1.5$ for SR in DLI and $p=1$ for QPR. Examples of an image in ImageNet and its pre-modulated versions with $p=1$ and $p=1.5$ are shown in Fig.\ref{fig:spm}. 

\begin{figure}[h!]
\centering\includegraphics[width=1\linewidth]{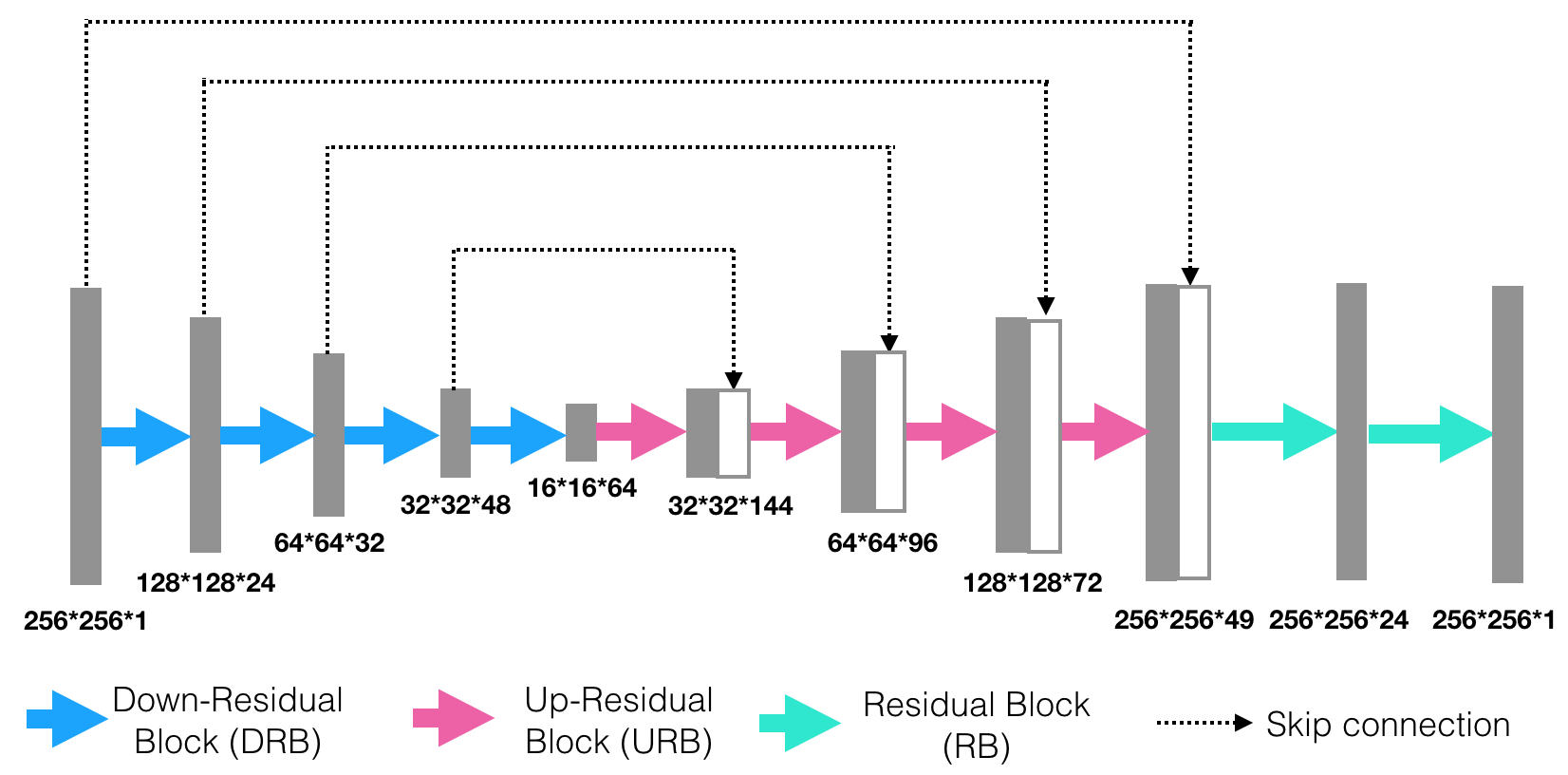}
\caption{Residual U-Net.}
\label{fig:PHENN}
\end{figure}

\subsection{Architectures of Deep Neural Networks}
DNN-L and DNN-H follow the same U-Net architecture \cite{ronneberger2015u} and utilize residuals to facilitate information flow  \cite{he2016deep}. As shown in Fig.\ref{fig:PHENN}, the input to both neural networks is the measurement $g$ with size 256x256x1, which is subsequently passed through four successive down-residual blocks (DRBs) for feature extraction. The extracted features pass through 4 up-residual blocks (URBs) and two residual blocks (RBs) for pixel-wise regression to size 256x256x1. Details of architectures of DRBs, URBs, and RBs can be found in the Supplement (Fig.A1). DNN-L is trained with the un-modulated ground truth $f$ while DNN-H is trained with the frequency pre-modulated version of the ground truth $\tilde{f}$, as described in Section \ref{sec:modu}. 

DNN-S takes in $\hat{f}^{LF}$ and $\hat{f}^{HF}$ as inputs. To best preserve the high frequency information contained in $\hat{f}^{HF}$, we do \textbf{not} have it pass through most layers of DNN-S. Instead, it only adds to the final outcome of $\hat{f}^{LF}$ passing through the Residual U-Net (Fig.\ref{fig:PHENN}) to form the final reconstruction $\hat{f}$. In other words, the DNN-S trains to map $\hat{f}^{LF}$ to a variant of itself that removes artifacts brought up by $\hat{f}^{HF}$ and (conceptually) synthesizes the low-frequency components contained in $\hat{f}^{LF}$ and the high-frequency components contained in $\hat{f}^{HF}$. A potential variation of DNN-S is to make it structurally the same as DNN-L (and DNN-H), except for the input to be instead the concatenation of $\hat{f}^{LF}$ and $\hat{f}^{HF}$(256x256x2) so that they both pass through all successive layers in the DNN. However, due to the low-frequency dominant nature of the DNN-S ground truth examples, exposure of $\hat{f}^{\text{HF}}$ to trainable convolutional filters may cause loss of high frequency information contained in $\hat{f}^{\text{HF}}$, as in DNN-L. This is why we opted for the first variant. 
\section{Results}
\label{sec:result}

\begin{figure*}[h!]
\centering\includegraphics[width=\linewidth]{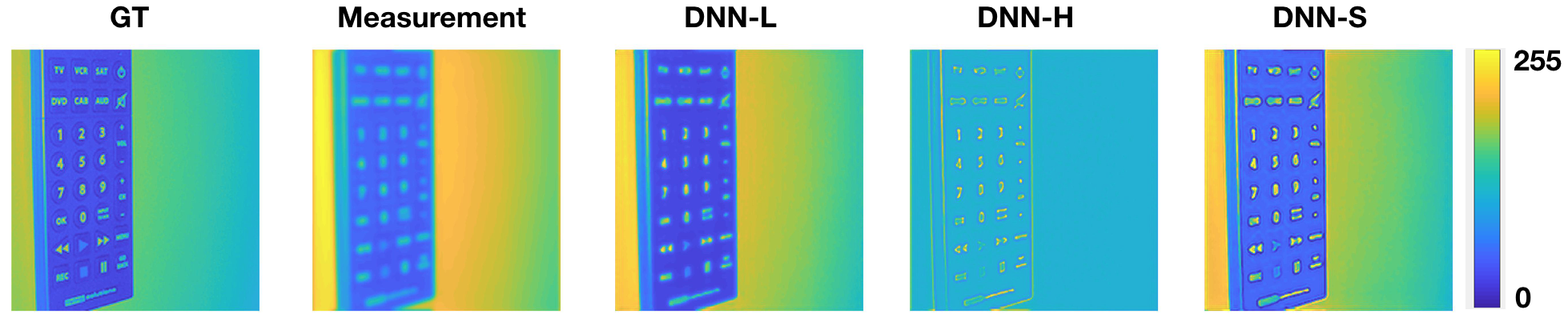}
\caption{Reconstruction results for SR in DLI (b=7).}
\label{fig:dli_spatial}
\end{figure*}

\subsection{Implementation details}

\paragraph{Training and Test Data} 
10,000 ImageNet\cite{russakovsky2015imagenet} images are used for training (9,000 for training and 1,000 for validation) and 100 ImageNet images and separately designed images for resolution test are used for testing. Measurement is obtained either synthetically (SR in DLI) or experimentally (QPR). For SR in DLI, the nominal resolving ability is set to be $b=7$ pixels. To apply the blur kernel properly, the ground truth is upsampled to $257\times 257$ before the blur kernel is applied and the outcome is then downsampled to $256\times 256$ to be the measurement $g$ used in DNN-L and DNN-H. In the case of QPR, $\lambda=0.633\mu\text{m}$ and $z=50\text{mm}$. In each training pair, the ground truth is a phase calibrated ImageNet image and the corresponding measurement is obtained by first subtracting a background pattern (also obtained experimentally) and then normalization, before being fed into the LS-DNN architecture. 

\paragraph{Negative Pearson Correlation Coefficient (NPCC) as the loss function}
We use the Negative Pearson Correlation Coefficient (NPCC) as the loss function to train all the neural networks mentioned above. If $k$ denotes the index in the training batch, and $(i,j)$ denotes the $(i,j)^{\text{th}}$ pixel in a particular image, then the batch-wise NPCC loss is defined as: 

\begin{equation}
\begin{aligned}
&\mathscr{L}_{\text{NPCC}}=\sum_{k}\mathscr{E}(f_{k},\hat{f_{k}}),\text{where},\\ 
&\mathscr{E}(f_{k},\hat{f_{k}})\equiv
\frac {-1\times\sum_{i,j}(f_{k}(i,j)-\overline{f_{k}})(\hat{f_{k}}(i,j)-\overline{\hat{f_{k}}})}{\sqrt{\sum_{i,j}(f_{k}(i,j)-\overline{f_k})^2}\sqrt{\sum_{i,j}(\hat{f_{k}}(i,j)-\overline{\hat{f_{k}}})^2}}.\\
\end{aligned}
\end{equation}
Here, $\overline{\cdot}$ denotes spatial averaging. Since NPCC only converges to the reconstruction within an affine transform, an additional histogram matching step \cite{li2018spectral} is performed.

\paragraph{Optimization}
The simulation is conducted on a Nvidia GTX1080 GPU using TensorFlow\cite{tensorflow2015-whitepaper}. Adam optimizer\cite{kingma2014adam} is used in the optimization of each deep neural network.  Each neural network is trained for 20 epochs and the batch size is 10.

\begin{figure*}[h!]
\centering\includegraphics[width=\linewidth]{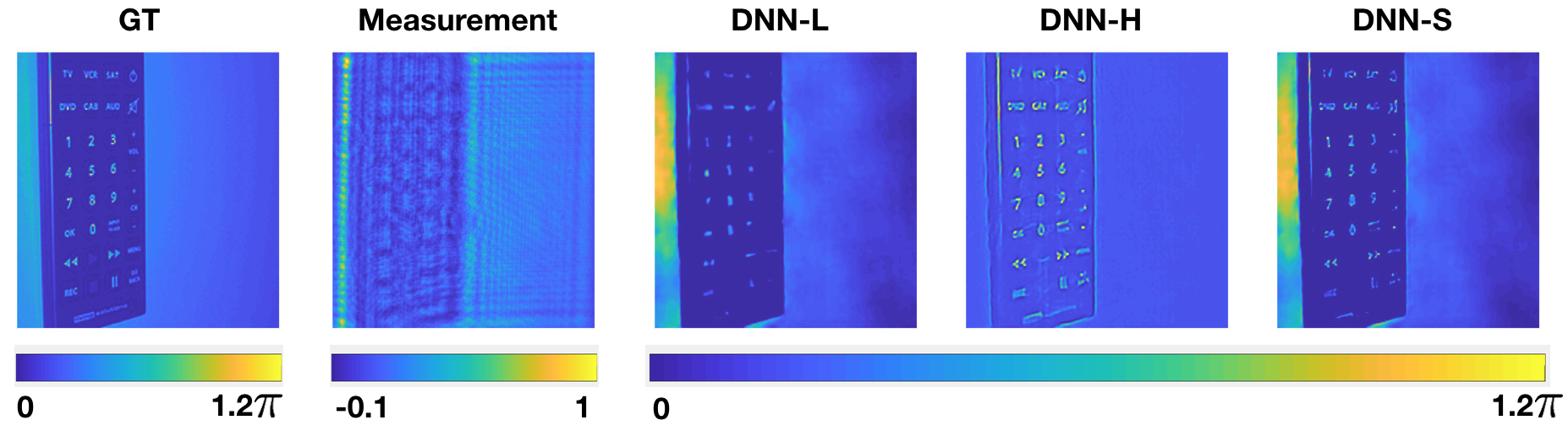}
\caption{Reconstruction results for QPR.}
\label{fig:phase_spatial}
\end{figure*}

\subsection{Reconstruction Results -- Spatial Domain}
Figs.\ref{fig:dli_spatial} and \ref{fig:phase_spatial} show test results for DLI and QPR, respectively. DNN-L, DNN-H and DNN-S label the outputs of the corresponding DNNs. In both scenarios, DNN-H succeeds in preserving high frequency components better than DNN-L, as expected, and the DNN-S removes the artifacts introduced by DNN-H through combining with the the low-frequency components captured well by DNN-L. More results are available in Supplement (Figs.A2,A3). Quantitative comparisons for the DNN-L and DNN-S outputs on the entire $100$ test images are shown in Table.\ref{tab:quant}. DNN-L actually has better quantitative performance in the problem of SR in DLI. This results verifies the well-known fact that neither PSNR nor SSIM is a good performance metric for visual quality\cite{gupta2011modified,wang2004image,wang2003multiscale,johnson2016perceptual,ledig2017photo}.

\begin{figure}[h!]
\centering\includegraphics[width=0.9\linewidth]{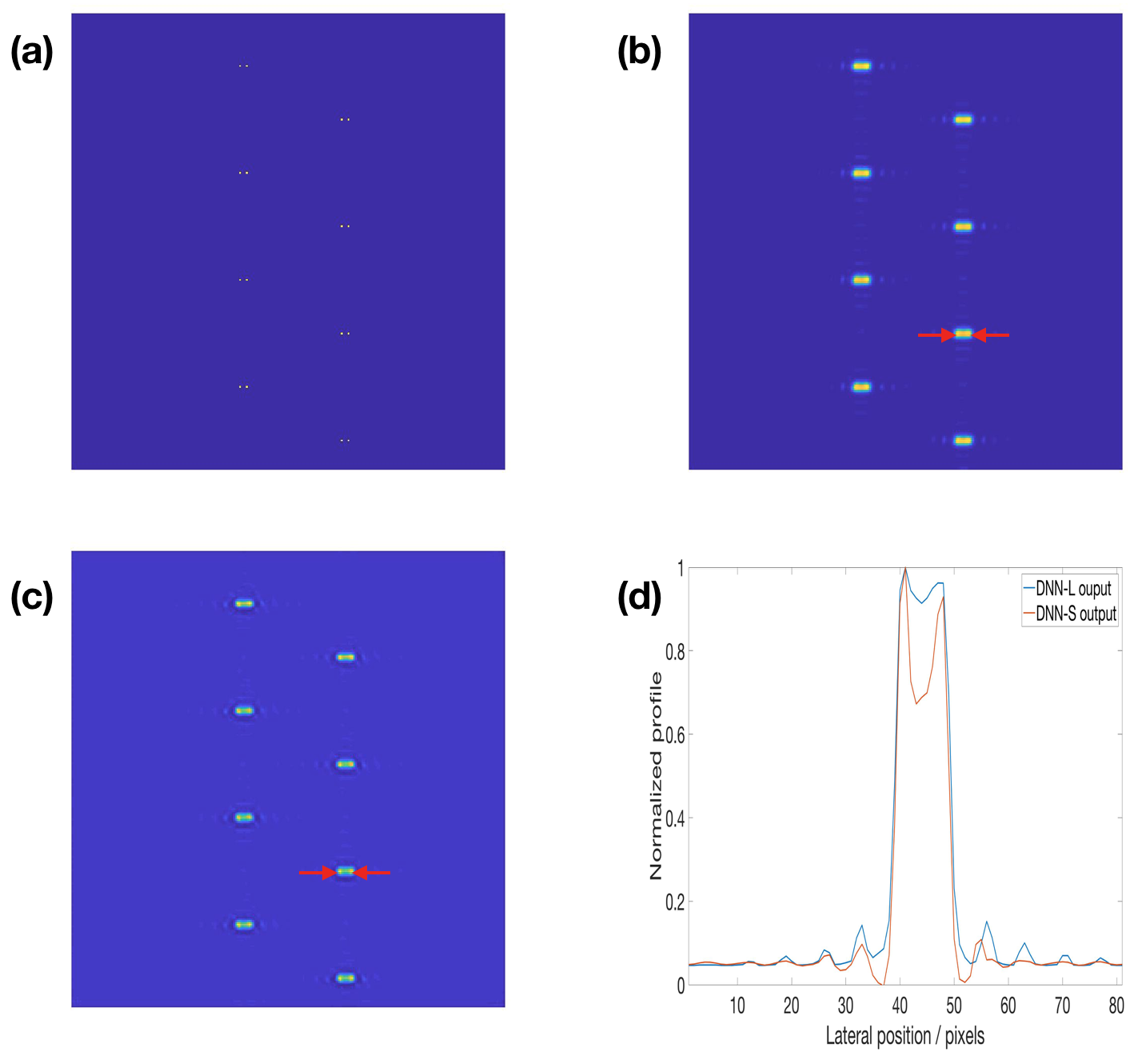}
\caption{Resolution test results for SR in DLI. (a) Dot pattern with spacing $D=5$ pixels, (b) DNN-L reconstruction, (c) DNN-S reconstruction, (d) 1D cross-sections along the line indicated by red arrows in (b) and (c).}
\label{fig:dli_res}
\end{figure}

Dot patterns are also used to test the resolution achieved by LS-DNN. The resolution test result for DLI is shown in Fig.\ref{fig:dli_res}. In the test pattern, the spacing $D$ between two nearby  dots is $5$ pixels, which is beyond the resolving ability of the DLI system ($7$ pixels). While the DNN-L reconstruction failed to distinguish the nearby dots, DNN-S output successfully resolves them. Even though the reconstructions in DNN-L already achieve super-resolution (Supplement Fig.A4), this result demonstrates that DNN-S improves the SR factor even further. The resolution test for phase retrieval is available in the Supplement (Fig.A5), and supports the same conclusion. 

To demonstrate that performance improvement of DNN-S over DNN-L is a not a sole consequence of computational capacity increase, we design another ResNet, DNN-L-3, which has the same architecture but twice as many feature maps (except in the last residual block) on each convolutional layer as its counterpart in DNN-L. The resulting DNN-L-3 has more than three times as many trainable parameters as DNN-L(or DNN-H, DNN-S). Reconstructions by DNN-L-3 are compared to those by DNN-S (Supplement Figs.A6, A7.) The superiority of DNN-S results proves our claim. 

\begin{table}[]
\caption{Quantitative evaluations of DNN-L and DNN-S performance in the two imaging scenarios.}
\centering
\begin{tabular}{@{}cll@{}}
\toprule
\multicolumn{1}{l}{}                                                   & \multicolumn{1}{c}{\begin{tabular}[c]{@{}c@{}}DNN-L\\ PSNR/SSIM\end{tabular}} & \multicolumn{1}{c}{\begin{tabular}[c]{@{}c@{}}DNN-S\\ PSNR/SSIM\end{tabular}} \\ \midrule
\begin{tabular}[c]{@{}c@{}}SR in DLI\end{tabular} & \textbf{22.9622/0.5545}                                                      & 19.9480/0.5210                                                               \\
\begin{tabular}[c]{@{}c@{}}QPR \end{tabular} & 18.7531/0.3684                                                               & \textbf{18.8525/0.3884}                                                      \\ \bottomrule
\end{tabular}
\label{tab:quant}
\end{table}

\subsection{Reconstruction Results -- Frequency Domain}
Next we analyze LS-DNN performance in the frequency domain. Figs.A8, A9 in the Supplement show the 2D Fourier spectra of the images in Figs.\ref{fig:dli_spatial} and \ref{fig:phase_spatial}, respectively. We find in both DLI and QPR scenarios that, as expected, while DNN-L works well in the low frequency range but fails to recover the high frequency components, DNN-H is better at retrieving the high frequency components but loses some low frequency information. DNN-S serves as a learned synthesizer of the DNN-L and DNN-H outputs at all frequency bands, low and high.

\begin{figure}[h!]
\centering\includegraphics[width=0.9\linewidth]{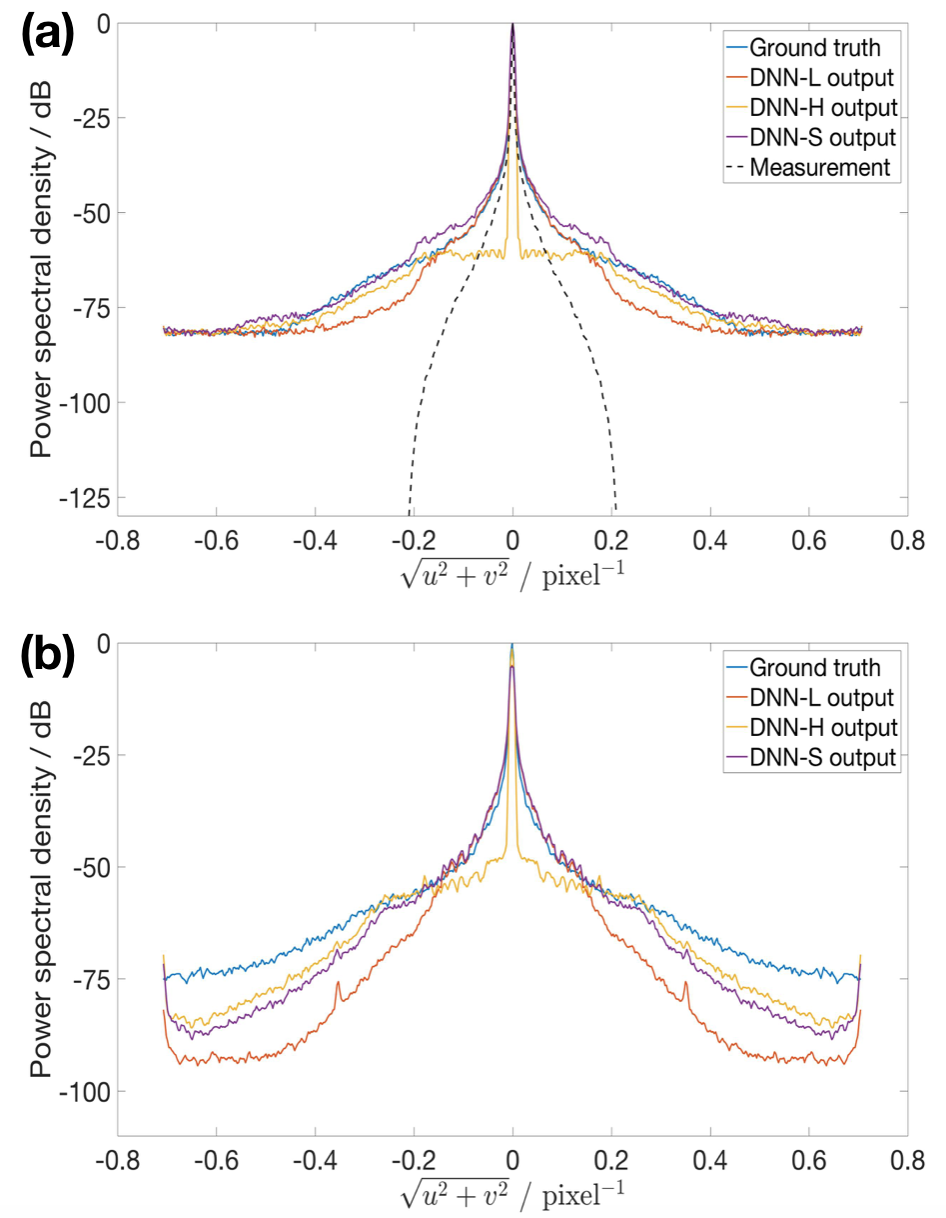}
\caption{1D cross-sections of the reconstructions' power spectral density (PSD) on $100$ ImageNet test images. (a) SR in DLI, (b) QPR.}
\label{fig:psd_1d}
\end{figure}

To investigate the performance of our approach on the entire test dataset, we compute the 2D power spectral density (PSD) for the reconstructions of the entire ensemble of the $100$ test images and show the 1D cross-sections (along the diagonal direction) in Fig.\ref{fig:psd_1d}. As expected, in both cases, the PSD for the DNN-L output matches well with the ground truth in the low frequency range but is much lower than the ground truth in the high frequency range. Notably, in the case of DLI, the point that the performance of DNN-L starts to drop almost coincides with the cutoff frequency in the measurement. The PSD for DNN-H output is getting closer to the ground truth in the high frequency range than the DNN-L output, benefiting from  training data pre-modulation. However, it is much worse in the low frequency range. The DNN-S output, which takes the advantage of both DNN-L and DNN-H, has a PSD that is close to the ground truth within the entire frequency range.  

\begin{figure*}[h!]
\centering\includegraphics[width=0.7\linewidth]{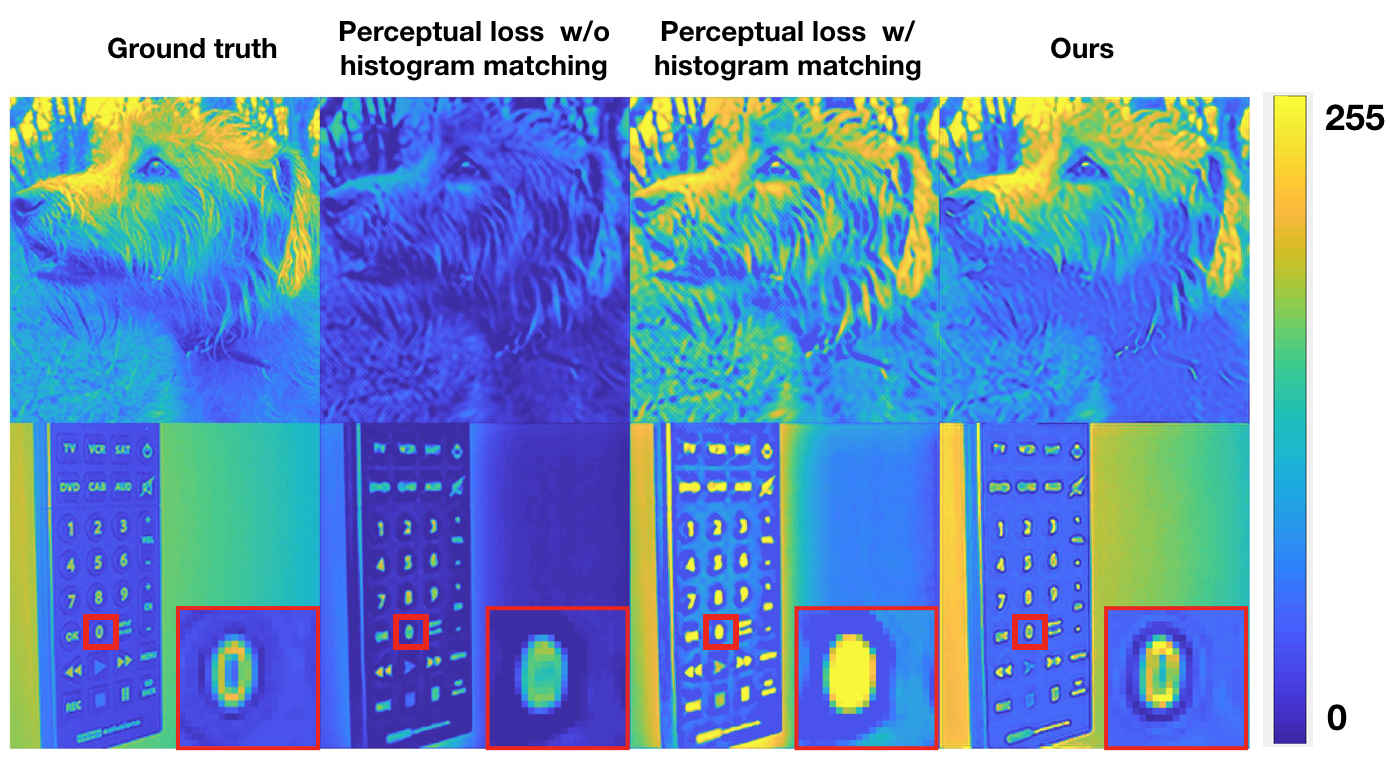}
\caption{Comparison with perceptual loss strategy \cite{johnson2016perceptual} for SR in DLI.}
\label{fig:perc_SR}
\end{figure*}

\begin{figure*}[h!]
\centering\includegraphics[width=0.65\linewidth]{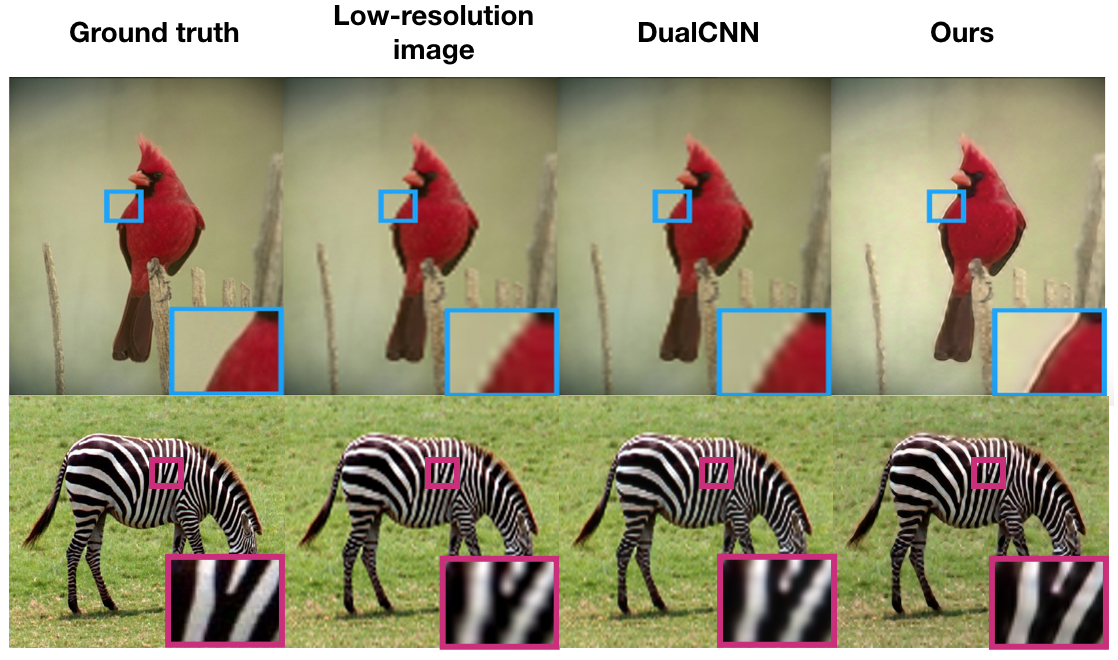}
\caption{Comparison with DualCNN \cite{Pan_2018_CVPR} for SR (4X).}
\label{fig:compare}
\end{figure*}

\subsection{Comparison with Perceptual Loss Strategy}
Methods with VGG \cite{simonyan2014very}-based perceptual loss \cite{johnson2016perceptual,ledig2017photo}, rather than pixel-wise (low-level), as the training loss function, have emerged as most promising approaches for fine detail enhancement in image reconstruction. We conduct a comparison of our DNN-S results with those obtained in \cite{johnson2016perceptual}, in the problem of SR in DLI. The DNN that we use for the perceptual loss strategy is DNN-L-3, so that the perceptual loss approach is not at a disadvantage in terms of  computational capacity. Due to the unclear description of the histogram matching step in \cite{johnson2016perceptual}, we present the results by perceptual loss strategy before and after a histogram matching step according to the widely used polynomial-fit scheme (Supplement Section S7 and Fig.A10). From Fig.\ref{fig:perc_SR}, our results are visually better than those by \cite{johnson2016perceptual}, with and without histogram matching. More comparison results available in Supplement Fig.A11. We also compare our results with those by using DNN-L for perceptual loss (Supplement Fig.A12). Our reconstructions perform better in the reconstruction of fine details, without adding undesired high-frequency artifacts.

\subsection{Comparison with DualCNN}
Lastly, we compare our approach with DualCNN \cite{Pan_2018_CVPR} for SR interpreted as upsampling, as used in \cite{Pan_2018_CVPR}. To ensure a fair comparison, we used the same dataset (BSDS500), upsampling ratio ($4\times$), pre-processing of data and neural network architectures for DNN-L and DNN-H as in \cite{Pan_2018_CVPR}. 

The comparison results are shown in Fig.\ref{fig:compare} and Table.\ref{tab:comp} (and more in Supplement Fig.A13), where visually better results are generated by our approach in Fig.\ref{fig:compare}, though Table.\ref{tab:comp} demonstrates DualCNN achieves a better quantitative performance, which is not the main objective of this work. This again verifies the well-known fact that neither PSNR nor SSIM is a good performance metric for visual quality\cite{gupta2011modified,wang2004image,wang2003multiscale,johnson2016perceptual,ledig2017photo}.   

\begin{table}[h!]
\caption{Quantitative comparison with DualCNN \cite{Pan_2018_CVPR} for SR (4X).}
\centering
\begin{tabular}{@{}lll@{}}
\toprule
                              & \multicolumn{1}{c}{Dual CNN} & \multicolumn{1}{c}{Ours} \\ \midrule
\multicolumn{1}{c}{PSNR/SSIM} & \textbf{24.5657/0.7953}      & 22.4155/0.7771           \\ \bottomrule
\end{tabular}
\label{tab:comp}
\end{table}

\section{Conclusions}
We have proposed a novel LS-DNN architecture that leverages the learning capability of a neural network to optimally synthesize parallel reconstructions that are reliable on low and high frequencies, respectively. The results verified our idea and showed reconstructions of better visual quality than the perceptual loss \cite{johnson2016perceptual} approach, as well as the more conceptually similar to ours DualCNN approach\cite{Pan_2018_CVPR}. 

The LS-DNN methodology is applicable to many other ill-posed image recovery problems, when there is imbalance between the spatial frequency region where the forward operator suppresses or loses content and the availability of content in the same frequency range in the training database to compensate. Generalizations may also be possible, for example to more than two frequency bands, similarly trained separately with pre-filtering and recombined in learned fashion; however, splitting the frequency bands too finely may be counterproductive if it results in too many parameters to learn. This is an interesting topic for future work. 



{\small
\bibliographystyle{ieee}
\bibliography{LS-DNN,inverse}
}

\end{document}